\documentclass{article}

  \usepackage[nonatpre]{neurips_2021}

\usepackage[utf8]{inputenc} 
\usepackage[T1]{fontenc}    
\usepackage{hyperref}       
\usepackage{url}            
\usepackage{booktabs}       
\usepackage{amsfonts}       
\usepackage{nicefrac}       
\usepackage{microtype}      
\usepackage{xcolor}         

\title{Rethinking Training from Scratch for \\ Object Detection}

\author{Yang Li, Hong Zhang, Yu Zhang\\
Zhejiang University\\
{\tt\small \{yanglivision, hongzhang99, zhangyu80\}@zju.edu.cn}
}

\begin{document}

\maketitle

\begin{abstract}
  The ImageNet pre-training initialization is the de-facto standard for object detection. He \etal \cite{RethinkImageNet} found it is possible to train detector from scratch(random initialization) while needing a longer training schedule with proper normalization technique. In this paper, we explore to directly pre-training on target dataset for object detection. Under this situation, we discover that the widely adopted large resizing strategy \eg resize image to (1333, 800) is important for fine-tuning but it's not necessary for pre-training. Specifically, we propose a new training pipeline for object detection that follows `pre-training and fine-tuning', utilizing low resolution images within target dataset to pre-training detector then load it to fine-tuning with high resolution images. With this strategy, we can use batch normalization(BN) with large bath size during pre-training, it's also memory efficient that we can apply it on machine with very limited GPU memory(11G). We call it \tb{direct detection pre-training}, and also use \tb{direct pre-training} for short. Experiment results show that direct pre-training accelerates the pre-training phase by more than 11x on COCO dataset while with even +1.8mAP compared to ImageNet pre-training. Besides, we found direct pre-training is also applicable to transformer based backbones \eg Swin Transformer. Code will be available at \href{https://github.com/wxzs5/direct-pretraining}{https://github.com/wxzs5/direct-pretraining}
\end{abstract}

\section{Introduction}

Recently, deep convolutional neural networks (CNNs) \cite{AlexNet,VGGNet, ResNet} made a remarkable progress for image classification. Then, \cite{RCNN, VisualizingNet} found that trained network from image classification can be well transferred to object detection training initialization. As time went on, numerous downstream vision tasks employed this `pre-training and fine-tuning' fashion. However, for downstream tasks, the pre-training is a time consuming work because many people believe that pre-training need massive size of datasets like ImageNet \cite{ImageNet} to get good or 'universal' visual representations. There are few works explored to train detectors from scratch, until He \etal \cite{RethinkImageNet} shows that on COCO \cite{COCO} dataset, it is possible to train comparably performance detector from scratch without ImageNet pre-training and also reveals that ImageNet pre-training speeds up convergence but can't improve final performance for detection task. But it still needs a relatively long training schedule \eg it totally catches up ImageNet pre-training needs 6x (72 epochs) schedule.

\begin{figure}
  \centering
  \includegraphics[width=0.48\linewidth]{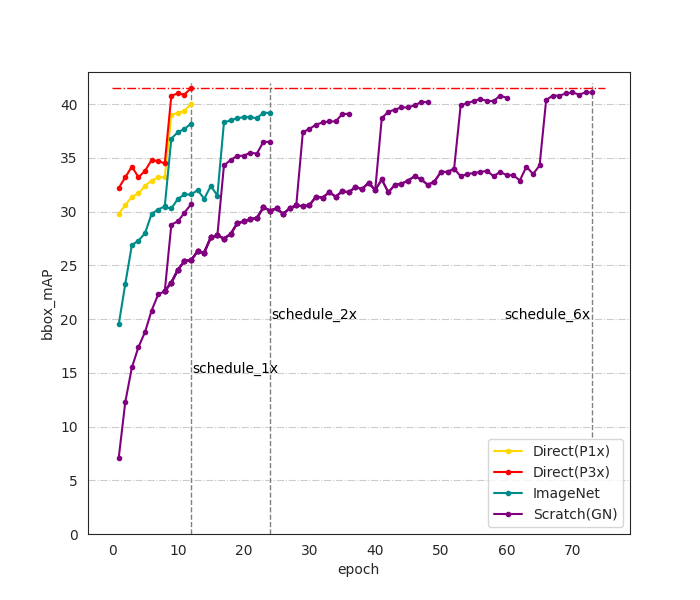}
  \includegraphics[width=0.48\linewidth]{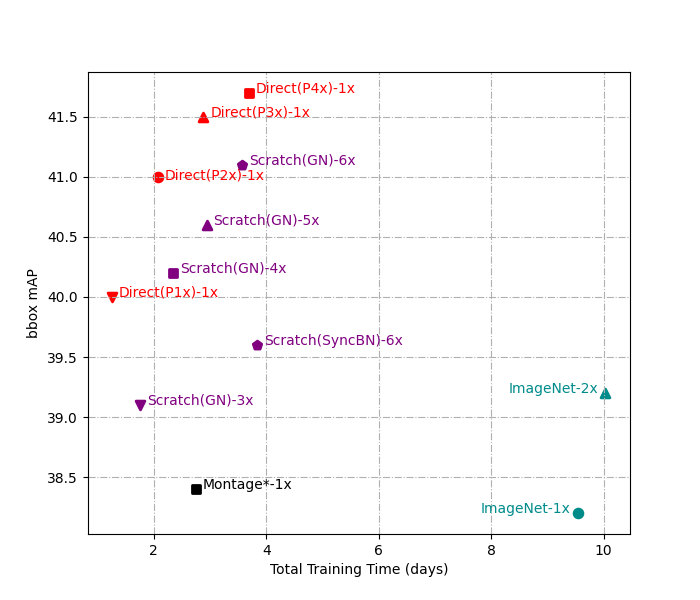}
  \begin{picture}(0,0)
    \put(-340,-8){(a) Fine-tuning performance}
    \put(-152,-8){(b) Overall time-AP comparison}
  \end{picture}
  \caption{Training time and performance comparison on COCO dataset. '1x','2x','6x' schedule for 12, 24 and  73 epochs respectively.'P1x' means pre-training with 1x schedule \ie 12 epochs. All models trained in a 8*2080Ti machine.}
  \label{process_comp}
\end{figure}

In this paper, we explore to setup a fast scratch training pipeline for object detection. To tackle this task, one important issue is \tb{batch size setting} during training. For accuracy consideration, existing detection training strategy always needs to re-scale images to a larger size \eg (1333, 800) to feed network. Under this resolution, detectors are always trained in small batch size(typically 2) situation due to the limited memory, and it severely degrades networks' performance. Previous solution \cite{RethinkImageNet} is using group normalization(GN)\cite{GN} or synchronized batch normalization(SyncBN)\cite{MegDet,PANet} to replace BN \cite{BN}. But on the other hand, GN increases model complexity and  memory consumption, SyncBN increases training time due to cross-GPU communication. So, can we just use BN with large batch size for object detection? Back to ImageNet pre-training pipeline, ImageNet pre-trained models usually feed with images resized to (224, 224) and it indeed speed up convergence for training detectors. And more importantly, with low resolution setting, typical ImageNet training batch size setting is quite larger compare to detection task. So, we follow ImageNet pre-training setting, using low resolution images to pre-train detector, then fine-tuning with larger resizing strategy.

By reviewing existing ImageNet pre-training pipeline and scratch training pipeline,that montage pre-training\cite{Montage} already demonstrated that following 'pre-training and fine-tuning' on target detection dataset is feasible to accelerate pre-training process. It only utilizes classification labels while leaves bounding boxes and mask labels unused. For efficiency consideration, if we fully utilize these 'strong' labels may be better for pre-training. On the other hand, their method is very complicate \eg it need crop instances and assemble multiple cropped instances to one image. And it also need a specific designed pre-training head(ERF Head) to train classification model through synthesized dataset. In this paper, direct pre-training pipeline doesn't introduce any new specific module. Through our detailed experiments and analysis in this paper, we found this idea has huge superiority versus existing methods.

Direct pre-training is a coarse-to-fine approach that it first train model on small-size images resized from target detection dataset just for better fine-tuning initialization. Experiments show it's effective and efficient. Specifically, direct pre-training has following advantages compared with existing strategies:\tb{1.} Compared to purely scratch solution, it is much simpler without introducing any new network modules(\eg GN or SyncBN) yet with higher accuracy, train and inference speed, see Fig \ref{process_comp} for details. \tb{2.} Compared to ImageNet/Montage pre-training solution, it can totally abandon the heavy pre-training process on ImageNet, which largely accelerate scratch detection training. It is illustrated in Fig \ref{instruction}. \tb{3.} It is also very memory efficient, \eg in this paper, we can apply direct pre-training in 8 2080Ti GPUs server(11G memory per GPU), while previous works always need large memory GPUs\cite{MegDet} or TPUs\cite{DropBlock}. We also provide a detailed empirical study about performance influence by batch size and resolution settings.

\section{Related Works}

\subsection{ImageNet Pre-training}
ImageNet-1k \cite{ImageNet} challenge is a large scale visual recognition task. Firstly, it was convolutional neural networks(CNNs) that bring huge performance gain to computer vision since AlexNet\cite{AlexNet} trained on ImageNet. Then developed Networks \cite{NetInNet, VGGNet, ResNet, ResNeXt, DenseNet} continuously achieve state-of-the-art performance with stronger feature representations. Besides, ImageNet pre-training are developed which utilize the trained networks on ImageNet for downstream target vision task as 'backbone initialization', \eg, detection \cite{RCNN}, segmentation \cite{FCN}, and pose-estimation \cite{Hourglass}, these tasks are indeed benefited a lot from pre-training. Recently, there are developed a batch of transformer based backbones\cite{ViT,DeiT,Swin} shows huge potential ability to replace CNN, and these models are also pre-trained on ImageNet-1k or even larger ImageNet-22k dataset.

\subsection{Object Detection}
R-CNN \cite{RCNN} first introduced CNN in detection task. Then Fast R-CNN \cite{FastRCNN} use CNN extracted features as input to localization and classification sub-tasks. Faster R-CNN \cite{FasterRCNN} proposed Region Proposal Network (RPN) to select potential RoI for further optimization. There is also another category of detectors called one-stage detector such as YOLO \cite{YOLO}, SSD \cite{SSD} which are mainly focused on real-time detection. Since Mask R-CNN \cite{MaskRCNN} simply extended \cite{FasterRCNN} with a mask head for instance segmentation. Object detection and instance segmentation are developed closer and closer. Subsequently, several methods \cite{PANet, HTC, MSRCNN} were developed following the pipeline of Mask R-CNN with some refinements of structural details to improve various aspects like multi-scale capability, alleviation of mismatch and the balance of data.

\subsection{Training from Scratch}
Fine-Tuning from ImageNet classification was already applied to R-CNN \cite{RCNN} and OverFeat \cite{OverFeat} very early. Following these results, most modern object detectors and many other computer vision algorithms employ this paradigm. DSOD \cite{DSOD} first reported a special designed one-stage object detector training from scratch. \cite{RethinkImageNet} reported results that train from scratch with a longer training schedule can be competitive to the ImageNet pre-training counterparts. \cite{PreTrainAnalysis} analyzed difference pre-training on object detection dataset and ImageNet pre-training. Montage pre-training \cite{Montage} is the first to synthesize on target detection dataset for efficient pre-training paradigm designed for object detection. \cite{RethinkPreAndSelf} take a further step, Pointing out that more detection data diminishes the value of ImageNet pre-training.

\section{Methodology}

\begin{figure*}
  \centering
  \includegraphics[width=\linewidth]{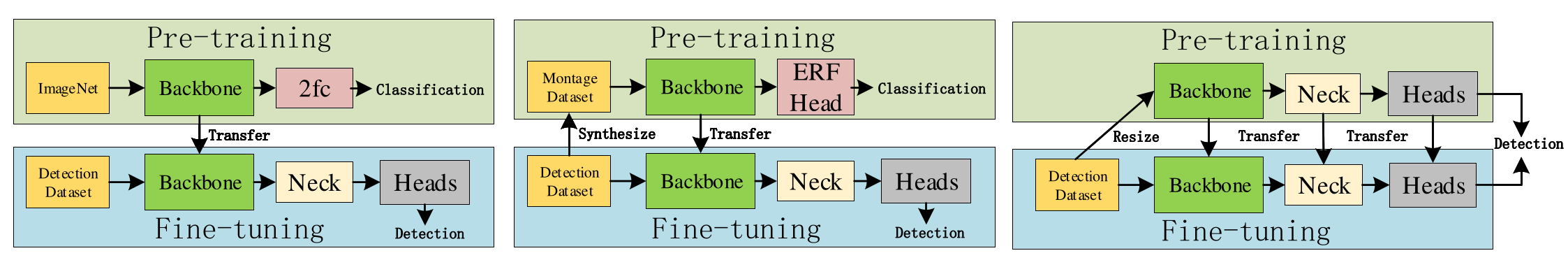}
  \begin{picture}(0,0)
    \put(-185,0){a) ImageNet Pre-training}
    \put(-65,0){b) Montage Pre-training \cite{Montage}}
    \put(60,0){c) Direct Detection Pre-training}
  \end{picture}
  \caption{Instruction of different pre-training paradigms. '2fc' in means two fully connected layers. 'Backbone' represents feature extraction Network, 'Neck' means FPN \etc 'Heads' for different RoI heads or dense heads.}
  \label{instruction}
\end{figure*}
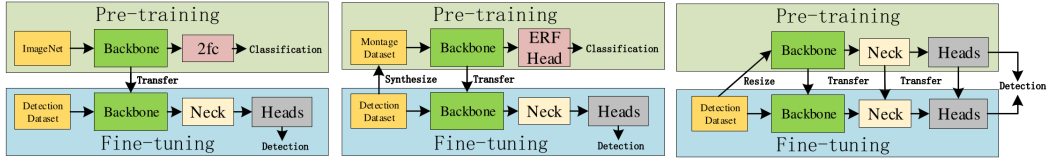

Our aim is to setup a fast training pipeline for object detection. To this end, we follow ImageNet pre-training paradigm to separate the training process into two phases:  pre-training phase and fine-tuning phase. Differently, for pre-training we directly utilize target detection dataset instead of classification dataset \eg ImageNet-1k. see Fig \ref{instruction} for different pre-training paradigms comparison. We first formulate problem in \S \ref{formulation}, and discuss design details in \S \ref{resize}, \S \ref{parameter}, \S \ref{norm}. More experiments setting details please read \S \ref{experiments}.

\subsection{Problem Formulation}\label{formulation}
Direct pre-training can be viewed as trade-off between batch size and input image resolution during pre-training. Specifically, we want to systematically investigate the performance influence when different image input size and batch size setting are applied. Besides, we need to take memory consumption into account in practice. Following \cite{EfficientNet}, A network can be defined as a function: $Y = F(X)$ where $F$ is network operation, $Y$ is output tensor, $X$ is input tensor, usually with tensor shape $\left\langle N, H, W, C\right\rangle$ during training. where $N$ is bath size, $H$ and $W$ are image spatial dimension and $C$ is the channel dimension. For typical RGB image, $C=3$ is a constant value. Setting a fixed steps $K$ for pre-training. In this paper, we follow previous work, view '1x schedule' is 12 epochs (88k steps). We can formulated this problem as:

\begin{equation}\label{problem}
  \begin{array}{ll}
    \max_{r, b}   & \operatorname{Accuracy}(\mathcal{F}(r, b))                                                            \\
    \text{ s.t. } & Y = \mathcal{F}(r, b) = \mathcal{F}(X_{\left\langle b \cdot N, r \cdot H, r \cdot W, C\right\rangle}) \\
                  & \operatorname{Memory}(\mathcal{F}(r, b)) \leq \text{target\_memory}
  \end{array}
\end{equation}

where r, b are coefficients for scaling input tensor during pre-training. It is obviously that both larger bath size and larger resolution would increase the memory consumption and training time. Let $\hat{N} = b \cdot N, \hat{H} = r \cdot H, \hat{W} = r \cdot W $. In practice, to fully utilize memory, we can manually set different $\hat{N}, \hat{H}, \hat{W}$ to make total memory consumption close to maximum GPU memory.

\subsection{Resizing Strategy}\label{resize}
Direct pre-training is so simple that it only changes a little in data pre-processing during pre-training, when fine-tuning it keeps all the same as typical data pre-processing does. Specifically, it only change the 'resize' process during data pre-processing, which typically is resized to (1333, 800). We reference existing two proposed resizing strategies: \tb{1.} ImageNet style, it resize input image to a fixed size \eg (256, 256), then randomly crop it to (224, 224). DropBlock\cite{DropBlock} also employed with this strategy. \tb{2.} Stitcher\cite{Stitcher} style, it firstly apply typical (1333, 800) resizing. Assume it resized to (H, W), then reduce image size by a divide factor $n$ to $(H/n, W/n)$. Both these strategies can reduce input image size which could leave more memory for setting larger bath size.

For ImageNet style resizing, due to randomly crop, it brings performance gain like typical multi-scale training which Stitcher style can't provide, we also experimentally compare the two different resizing strategy, under $batchsize=8$ Mask-RCNN with R50-FPN and the other factors keeps same,  ImageNet style got 40.0 mAP \vs Stitcher 39.4 mAP. Another disadvantage for Stitcher style is that due to random mini-batch training, there may appears memory consumption unstable during training, \eg a large image batched with a small image. While for ImageNet style, all image finally cropped to a fixed size \eg (224, 224), it's memory consumption is relatively stable, which is good for fully utilize GPU memory, and Stitcher style some times will cause memory overflow during training. Based on above consideration, we tend to use ImageNet style as resizing strategy.

Based on above analysis, we describe direct pre-training resizing strategy more detailedly: select a base image shape (L, L), randomly select a factor $\alpha \in (0.8, 1.2)$, then resize image to $(\alpha \cdot L, \alpha \cdot L)$. Finally randomly cropped or pad to fixed image shape (L, L).

\subsection{Hyper-parameter Setting}\label{parameter}
\begin{figure}
  \centering
  \includegraphics[width=\linewidth]{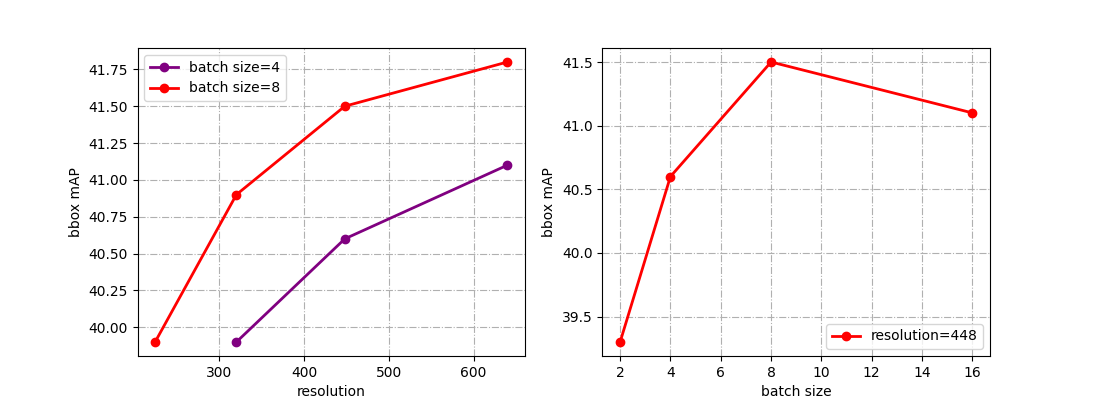}
  \caption{Batch and resolution setting performance comparison, all experiments sufficiently pre-training in 3x schedule(36 epochs, 261k iterations), and fine-tuning in 1x schedule.}
  \label{batch_reso}
\end{figure}

It's not easy to precisely solve the optimization problem equation \ref{problem}. In practice, we can utilize grid search to roughly find out better hyper-parameter setting. So we empirically set a batch of grid search experiments to observe the influence by batch size and input image resolution as a indirect solution. We set image size as square shape \eg (224, 224) for input, in experiments we set image length($\hat{H}$ or $\hat{W}$) as [224, 320, 448, 640] and batch size($\hat{N}$) as [2, 4, 8, 16] respectively, see results in Fig \ref{batch_reso}. We can observe that \tb{either increase resolution or batch size will improve final performance, and it will be saturated when these factors gradually increased}. From another perspective, when these factors increased, memory consumption and training time would also increase. Based on these observations from table \ref{batch}, we set direct pre-training baseline with $batchsize=8, resolution=448$ or in other words $\hat{N}=8, \hat{H}=448, \hat{W}=448$ in this paper.

From table \ref{batch}, it's worth to notice that even with $batchsize=2, resolution=448$ extreme situation, it still got final performance 39.3 mAP which already outperforms ImageNet fine-tuning 2x counterpart with 39.2 mAP. For comparison, scratch training with BN in 4x schedule only got 36.1 mAP. With above \S \ref{resize} analysis, it indicates that \tb{the random crop operation brings a lot gain to network}. And this technique widely used in ImageNet pre-training, this may provide another perspective to explain why ImageNet pre-training with very low resolution images (224, 224) but still could help these very large resizing tasks like object detection(1333, 800).

\begin{table}
  \centering
  \setlength{\tabcolsep}{3mm}{
    \begin{tabular}{c|c|c|c|c|cc}
      \toprule
      Method                & Batch            & Input                     & Schedule              & Cost (days) & $AP^{bbox}$ & $AP^{mask}$ \\
      \hline
      \multirow{3}*{Direct} & 2                & \multirow{3}*{(448, 448)} & \multirow{3}*{P3x+1x} & \tb{0.84}   & 39.3        & 35.1        \\
      ~                     & 4                & ~                         & ~                     & 1.37        & 40.6        & 36.3        \\
      ~                     & 8                & ~                         & ~                     & 1.76        & \tb{41.5}   & \tb{37.0}   \\
      \hline
      \multirow{3}*{Direct} & \multirow{3}*{4} & (320, 320)                & \multirow{3}*{P3x+1x} & \tb{1.11}   & 39.9        & 35.7        \\
      ~                     & ~                & (448, 448)                & ~                     & 1.37        & 40.6        & 36.3        \\
      ~                     & ~                & (640, 640)                & ~                     & 1.76        & \tb{41.1}   & \tb{36.8}   \\
      \hline
      ImageNet              & \multirow{3}*{2} & (1333, 800)               & 1x                    & -           & 38.2        & 34.7        \\
      Scratch(BN)           & ~                & (1333, 800)               & 4x                    & 2.16        & 36.1        & 32.4        \\
      Direct                & ~                & (448, 448)                & P3x+1x                & \tb{0.84}   & \tb{39.3}   & \tb{35.1}   \\
      \bottomrule
    \end{tabular}}
  \caption{Batch size and resolution setting performance comparison. 'P3x+1x' means pre-training in 3x and fine-tuning with 1x. }
  \label{batch}
\end{table}

\subsection{Normalization}\label{norm}

In this section, we separately discuss normalization during pre-training and fine-tuning. For pre-training, we use normal BN\cite{BN}, and due to the different batch size setting, we follow \tb{linear scaling rule}\cite{LargeSGD} to set learning rate, we use typical ImageNet fine-tuning learning rate 0.02 as base learning rate for $batchsize=2$, then when the batch size is multiplied by k, multiply the learning rate by k. Following batch normalization definition\cite{BN}, BN can be formulated as:

\begin{equation}
  y=\frac{x-E[x]}{\sqrt{\operatorname{Var}[x]+\epsilon}} * \gamma+\beta
  \label{BN}
\end{equation}

Input features is $x$, and output features is $y$, $\epsilon$ is a small value to avoid zero divide error. BN operation need calculate batch statistics mean $E[x]$ and variance $Var[x]$, and the $\gamma$, $\beta$ are learnable parameters. During fine-tuning, we observed that the normalization technique setting could significantly influence final performance. Existing solution is using pre-trained batch statistics as fixed parameters\cite{ResNet}, while it still leaves that learnable affine parameters $\gamma$, $\beta$ activated for updating when fine-tuning, we call this 'affine' style BN. We also call that totally fix all BN parameters during fine-tuning 'Fixed' style, and synchronized batch normalization is 'SyncBN' style. We systematically investigate the performance influence of these BN setting strategies, see table \ref{affine} for details. From it we find that totally fixed BN is the best, which even better than SyncBN style. Besides, it also makes the training process accelerates roughly 0.06s/iteration. And typical used 'affine' style BN would significantly degrade performance(40.4mAP to 37.8mAP). On the contrary, if we use fixed BN parameters for ImageNet fine-tuning it would slightly degrade performance compare to affine solution.

\begin{table}
  \centering
  \setlength{\tabcolsep}{1.5mm}{
    \begin{tabular}{c|c|ccc|c|cc}
      \toprule
      Pre-training Model         & Schedule          & Affine  & SyncBN  & Fixed   & Time(s/iter) & $AP^{bbox}$ & $AP^{mask}$ \\
      \hline
      \multirow{2}*{ImageNet}    & \multirow{2}*{2x} & $\surd$ &         &         & 0.479        & \tb{39.2}   & \tb{35.4}   \\
      ~                          & ~                 &         &         & $\surd$ & \tb{0.424}   & 38.7        & 35.1        \\
      \hline
      \multirow{3}*{Direct(P1x)} & \multirow{3}*{2x} & $\surd$ &         &         & 0.479        & 37.8        & 33.9        \\
      ~                          & ~                 &         & $\surd$ &         & 0.483        & 37.9        & 34.0        \\
      ~                          & ~                 &         &         & $\surd$ & \tb{0.424}   & \tb{40.4}   & \tb{36.1}   \\
      \bottomrule
      \hline
    \end{tabular}}
  \caption{Influence of fine-tuning BN strategies. 'Time' is training time cost per iteration during fine-tuning}
  \label{affine}
\end{table}

\section{Experiments}

\subsection{Experimental Settings}\label{experiments}

\tb{Dataset.} In this paper, if without specification, we evaluate our approach on MS-COCO \cite{COCO} datasets. We train the models on the COCO \textsl{train2017} split, and evaluate on COCO \textsl{val2017} split. We evaluate bbox Average Precision (AP) for object detection and mask AP for instance segmentation.

\tb{Implementation.} We implement our algorithm based on MMDetection\cite{MMDetection}. In the experiments, our model is trained on a server with 8 RTX 2080Ti GPUs. Our baseline is the widely adopted Mask R-CNN \cite{MaskRCNN} with FPN \cite{FPN} and backbone ResNet-50 \cite{ResNet}. Following \cite{RethinkImageNet} scratch training, there is a little difference compared with Mask R-CNN that bbox head is a \textsl{4conv1fc} head instead of \textsl{2fc} head. Originally, our pretrained ImageNet ResNet-50 backbone is from torchvision. Actually, these models are given in MMDetection's MODELZOO, and we use the 'pytorch' style in backbone setting. During fine-tuning, if without specification, we directly load whole weights of pre-trained Mask R-CNN for initialization, \ie we load backbone, neck(FPN), RPN and heads from pre-trained model.

\tb{Scheduling.} Following MMDetection setting, scheduling is based on epoch, 1x for 12 epochs. Pre-training schedule use fixed learning rate which dose not reduce learning rate while fine-tuning will follow typical setting to reduce learning rate.

\tb{Hyper-parameter.} All other hyper-parameters follow those in MMDetection\cite{MMDetection}. Specially, during pre-training, we set the learning rate as 0.08 for baseline with a batch size of 8.With regard to fine-tuning, all the setting is the same as baseline MMDetection R50-FPN.

\subsection{Main Results}

\begin{table*}
  \centering
  \setlength{\tabcolsep}{2.5mm}{
    \begin{tabular}{c|c|cc|cc}
      \toprule
      Method                 & Schedule & Pre-cost(days)          & Total Cost(days)        & $AP^{bbox}$          & $AP^{mask}$          \\
      \hline
      ImageNet\cite{Montage} & 1x       & 6.80                    & -                       & 37.3                 & -                    \\
      Montage\cite{Montage}  & 1x       & 1.73{\fo(-74.5\%)}      & -                       & 37.5{\fo(+0.2)}      & -                    \\
      \hline
      ImageNet               & 1x       & 9.05                    & 9.54                    & 38.2                 & 34.7                 \\
      Montage$\ast$          & 1x       & 2.26{\fo(-74.5\%)}      & 2.75{\fo(-71.2\%)}      & 38.4{\fo(+0.2)}      & -                    \\
      Direct(P1x)            & 1x       & \tb{0.81{\fo(-90.9\%)}} & \tb{1.26{\fo(-86.7\%)}} & 40.0{\fo(+1.8)}      & 35.8{\fo(+1.1)}      \\
      Direct(P2x)            & 1x       & 1.62{\fo(-92.0\%)}      & 2.07{\fo(-87.7\%)}      & \tb{41.0{\fo(+2.8)}} & \tb{36.6{\fo(+1.9)}} \\
      \hline
      ImageNet               & 2x       & 9.05                    & 10.03                   & 39.2                 & 35.4                 \\
      Scratch(SyncBN)        & 6x       & -                       & 3.84{\fo(-56.8\%)}      & 39.6{\fo(+0.4)}      & 35.5{\fo(+0.1)}      \\
      Scratch(GN)            & 6x       & -                       & 2.35{\fo(-71.7\%)}      & 41.1{\fo(+1.9)}      & 36.5{\fo(+1.1)}      \\
      Direct(P3x)            & 1x       & \tb{1.07{\fo(-88.2\%)}} & \tb{1.52{\fo(-84.1\%)}} & 41.5{\fo(+2.3)}      & 37.0{\fo(+1.6)}      \\
      Direct(P4x)            & 1x       & 1.44{\fo(-84.1\%)}      & 1.89{\fo(-80.1\%)}      & \tb{41.7{\fo(+2.5)}} & \tb{37.3{\fo(+1.9)}} \\
      \bottomrule
    \end{tabular}}
  \caption{Main Mask R-CNN results on COCO dataset. P2x,P3x,P4x means pre-training in 24, 36 and 48 epochs respectively.  For better comparison, Montage* are setting from reported values 'normalize' version. Note: Montage time cost from V100 GPU, while ours comes from 2080Ti and it's speed is about 0.79x V100. }
  \label{total_res}
\end{table*}

In this section, we compare our models with typical 1x and 2x settings. Results shows in table \ref{total_res}, also shows in Fig \ref{total_res}. For Montage pre-training, we use it originally reported results\cite{Montage}, for fair comparison, we 'normalize' its' performance and cost time by add to our ImageNet baseline. Following \cite{Montage}, All reported cost time is only account GPU running time which ignores data pre-processing time. Actually direct pre-training has faster data pre-training time see \S \ref{speed} for details. Then we calculate the Cost of time via their papers' reported time consumption to compare with ImageNet pre-training. We can see that our models are significantly fast and accurate compared to ImageNet ones. Direct(P1x) model is already with 1.8 mAP higher than typical ImageNet 1x model, further more, it outperforms ImageNet 2x model. And we compare it's pre-training phase to ImageNet one, it accelerate more than 11x(-90.9\% GPU days).

For longer schedule \eg Scratch(GN)-6x and Scratch(SyncBN)-6x, we can see direct pre-training is still with higher efficiency from Fig \ref{total_res} (b). It is worth to note that Scratch(SyncBN) has the same network architecture, but with different training strategy, even in 6x schedule which is sufficiently trained, it still has large performance margin compare to direct pre-training in 4x schedule(39.6mAP \vs 41.7 mAP).

\subsection{Schedule Setting Trade-off}

\begin{figure}
  \begin{floatrow}[2]
    \tablebox{\caption{Different schedule setting model performance comparison.}\label{trade}}{
      \setlength{\tabcolsep}{1mm}{
        \begin{tabular}{c|c|cc}
          \toprule
          Method                     & Schedule & $AP^{bbox}$ & $AP^{mask}$ \\
          \hline
          \multirow{2}*{Direct(P1x)} & 1x       & 40.0        & 35.8        \\
          ~                          & 2x       & \tb{40.4}   & \tb{36.1}   \\
          \hline
          \multirow{2}*{Direct(P2x)} & 1x       & \tb{41.0}   & \tb{36.6}   \\
          ~                          & 2x       & 40.8        & 36.4        \\
          \hline
          \multirow{2}*{Direct(P3x)} & 1x       & \tb{41.5}   & \tb{37.0}   \\
          ~                          & 2x       & 41.0        & 36.6        \\
          \hline
          Direct(P4x)                & 1x       & \tb{41.7}   & \tb{37.3}   \\
          \bottomrule
        \end{tabular}}}
    \figurebox{\caption{Fine-tuning performance trade-off on COCO dataset.}\label{finetune}}{
      \includegraphics[width=0.7\linewidth]{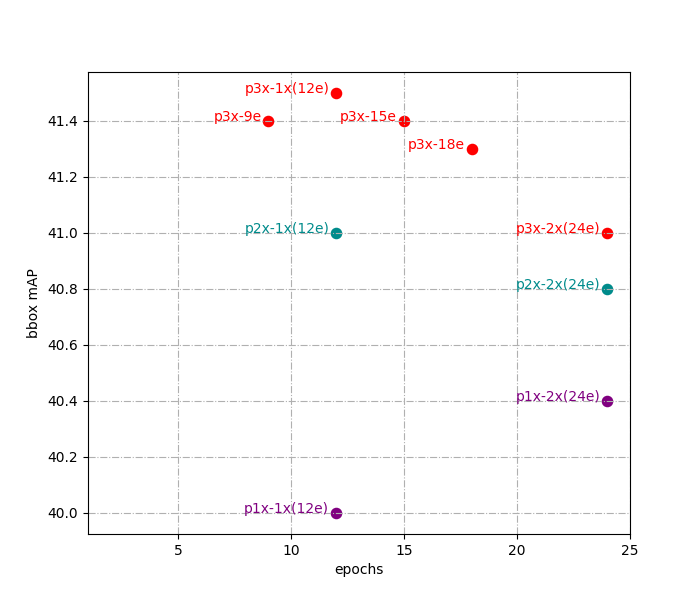}}
  \end{floatrow}
\end{figure}

We investigate the influences of schedule setting on pre-training and fine-tuning respectively in this part. We observe that increasing pre-training iterations will provide better pre-trained models, and the performance upper bound are mainly determined by pre-training instead of fine-tuning. We set fine-tuning schedule fixed in 1x schedule and pre-training schedule variate to find out the performance influence. Results are shown in table \ref{trade}. Results shows that in 3x schedule for pre-training is sufficiently enough. Then we investigate fine-tuning schedule settings in Fig \ref{finetune}. We can observe that direct pre-training is easily to over-fitted. The longer pre-training schedule adopted, the more easily for fine-tuning to get over-fitted. Fig \ref{finetune} shows when pre-training in 3x schedule, best fine-tuning schedule should between 9 epochs and 15 epochs. So, summarize above observations, we suggest that \tb{pre-training in 3x schedule and fine-tuning with 1x schedule is sufficiently trained for direct pre-training}.

\subsection{Gain Analysis}

Different from typical ImageNet pre-training, direct pre-training includes new model parts(neck, RPN and RoI Heads) during pre-training instead of just backbone. So in this section, we quantize the performance influence of different model parts applied with direct pre-training, to find out whether the increased model parts help training. Results shows in table \ref{gain}. It proved these parts indeed improve the detection performance. Thereinto, bounding box head helps most while RPN helps least even slightly degrade performance(-0.1mAP).

\begin{table}
  \centering
  \setlength{\tabcolsep}{5mm}{
    \begin{tabular}{c|cc}
      \toprule
      Load Parameters & $AP^{bbox}$          & $AP^{mask}$          \\
      \hline
      ImageNet        & 38.2                 & 34.7                 \\
      \hline
      Backbone        & 39.0                 & 34.7                 \\
      + FPN           & 39.4{\fo(+0.4)}      & 35.2{\fo(+0.5)}      \\
      + RPN           & 39.3{\fo(-0.1)}      & 35.1{\fo(-0.1)}      \\
      + (Bbox Head)   & \tb{41.7{\fo(+2.4)}} & 36.8\tb{\fo(+1.7)}   \\
      + (Mask Head)   & \tb{41.7}{\fo(+0.0)} & \tb{37.3}{\fo(+0.5)} \\
      \bottomrule
    \end{tabular}}
  \caption{Gain analysis over different model parts from direct pre-training. }
  \label{gain}
\end{table}

\subsection{Multi-scale Training}

Due to the crop operation during pre-training, which can be viewed as a multi-scale training trick. We need to answer the question: if we use other pre-training strategy, can it catch up direct pre-training performance when adopt typical multi-scale training in a longer fine-tuning schedule? We use direct(P1x) as baseline, to compare with ImageNet and Scratch(SyncBN) counterparts trained with multi-scale training. we keeps fine-tuning steps the same, apply with poly multi-scale strategy which randomly resize short image edge to [640, 672, 704, 736, 768, 800] and evaluate performance, results shows in table \ref{multi_scale}. We can see that direct pre-training still has superiority over counterparts, even direct pre-training in 2x schedule without applying multi-scale training, it still outperforms ImageNet with multi-scale training.(40.4mAP \vs 40.2mAP)

\begin{table}
  \centering
  \setlength{\tabcolsep}{3mm}{
    \begin{tabular}{c|c|c|cc}
      \toprule
      Method          & Schedule & Multi-Scale & $AP^{bbox}$ & $AP^{mask}$ \\
      \hline
      ImageNet        & 2x       &             & 39.2        & 35.4        \\
      Direct(P1x)     & 1x       &             & 40.0        & 35.8        \\
      Direct(P1x)     & 2x       &             & \tb{40.4}   & \tb{36.1}   \\
      \hline
      ImageNet        & 2x       & $\surd$     & 40.2        & 36.1        \\
      Scratch(SyncBN) & 3x       & $\surd$     & 36.0        & 32.8        \\
      Direct(P1x)     & 2x       & $\surd$     & \tb{41.0}   & \tb{36.7}   \\
      \bottomrule
    \end{tabular}}
  \caption{Multi-scale training performance comparison. }
  \label{multi_scale}
\end{table}

\subsection{Train and Inference Time Comparison}\label{speed}

Our method is faster than typical counterparts ImageNet and scratch(GN) during fine-tuning. During inference, speed of our method is actually the same as ImageNet fine-tuning models. When it comes to scratch(GN) training, without the introduction of GN, our model is faster during both fine-tuning and inference. Results are shown in table \ref{time}. We get train time and data pre-processing time by averaging all training steps of P3x model during fine-tuning on 8 GPUs. And inference speed is calculated from inference COCO val2017 on a 1080Ti GPU.

\begin{table}
  \centering
  \setlength{\tabcolsep}{2mm}{
    \begin{tabular}{c|c|c|c}
      \toprule
      Method          & Time (s/iter) & Inference (fps) & Data Time (s) \\
      \hline
      ImageNet        & 0.478         & \tb{8.6}        & -             \\
      Scratch(GN)     & 0.578         & 7.6             & 0.029         \\
      Scratch(SyncBN) & 0.483         & \tb{8.6}        & 0.029         \\
      Direct          & \tb{0.424}    & \tb{8.6}        & \tb{0.014}    \\
      \bottomrule
    \end{tabular}}
  \caption{Different model time computation comparison. 'Time' is training time cost per iteration during fine-tuning,'Data Time' is data pre-processing time.}
  \label{time}
\end{table}

\subsection{More Validations}

In this section we validate effectiveness of direct pre-training on more detectors. To this end, we select typical one-stage, two-stage and multi-stage detectors to validate, different from our Mask-RCNN baseline, these detectors trained without mask label. While we see results in table \ref{more}, all these detectors applied with direct pre-training outperform ImageNet counterparts. Further more, we also validate it on recently processed transformer based backbone Swin Transformer \cite{Swin}, we use Swin-T backbone,and do not fix layer normalization(LN) layer in backbone we keeps in activated. results shown in table \ref{more}, it proves that \tb{direct pre-training still worked for transformer based backbones.}

\begin{table}
  \centering
  \setlength{\tabcolsep}{2mm}{
    \begin{tabular}{c|c|cccccc}
      \toprule
      Method                           & Pre-training & AP        & $AP_{50}$ & $AP_{75}$ & $AP_s$    & $AP_m$    & $AP_l$    \\
      \hline
      \multirow{2}*{RetinaNet}         & ImageNet     & 36.5      & 55.4      & 39.1      & 20.4      & \tb{40.3} & 48.1      \\
      ~                                & Direct(P1x)  & \tb{37.1} & \tb{55.7} & \tb{39.6} & \tb{22.1} & 40.0      & \tb{48.6} \\
      \hline
      \multirow{2}*{Faster RCNN}       & ImageNet     & 37.4      & 58.1      & 40.4      & 21.2      & 41.0      & 48.1      \\
      ~                                & Direct(P1x)  & \tb{39.3} & \tb{58.7} & \tb{43.0} & \tb{24.4} & \tb{42.1} & \tb{49.8} \\
      \hline
      \multirow{2}*{Cascade RCNN}      & ImageNet     & 40.3      & 58.6      & 44.0      & 22.5      & 43.8      & 52.9      \\
      ~                                & Direct(P1x)  & \tb{41.5} & \tb{59.4} & \tb{45.1} & \tb{25.3} & \tb{44.2} & \tb{53.4} \\
      \hline
      \multirow{2}*{Mask RCNN w/ Swin} & ImageNet     & 43.8      & 65.3      & 48.1      & 26.8      & 47.4      & \tb{58.0} \\
      ~                                & Direct(P1x)  & \tb{45.0} & \tb{65.6} & \tb{49.6} & \tb{30.2} & \tb{48.2} & 57.4      \\
      \bottomrule
    \end{tabular}}
  \caption{Different models direct pre-training performance comparison.}
  \label{more}
\end{table}

\subsection{Direct Pre-training with Less Data}\label{lessdata}
In this section we validate our method's effectiveness with less data situation. We validate on 1/10 of COCO dataset(11.8k images) and PASCAL VOC \cite{VOC} dataset. On 1/10 COCO, we trained it with roughly 60k steps and 120k steps for direct pre-training \vs fine-tuning 60k with ImageNet pre-training. Hyper-parameter are follows \cite{RethinkImageNet} \eg base learning rate is set to 0.04 and decay factor is 0.02. Differently, we do not use multi-scale for consistency with above settings. For PASCAL VOC we train a Faster R-CNN with \textsl{train2007+train2012} and evaluate on \textsl{val2007}. Results are shown in table \ref{less}. From this results, under 1/10 COCO dataset direct pre-training still outperforms ImageNet pre-training counterpart, but for PASCAL VOC direct pre-training can't catch up with ImageNet pre-training. It shows that \tb{in extremely less data situation like PASCAL VOC dataset direct pre-training still can't replace ImageNet pre-training.}

\begin{table}
  \centering
  \setlength{\tabcolsep}{4mm}{
    \begin{tabular}{c|c|c|c|c}
      \toprule
      Dataset                  & Model                      & Method   & Iterations & $AP^{bbox}$ \\
      \hline
      \multirow{2}*{1/10 COCO} & \multirow{2}*{Mask R-CNN}  & ImageNet & 60k        & 22.1        \\
      ~                        & ~                          & Direct   & P60k+30k   & \tb{23.2}   \\
      \hline
      \multirow{2}*{VOC}       & \multirow{2}*{Faster RCNN} & ImageNet & 12.4k      & \tb{79.47}  \\
      ~                        & ~                          & Direct   & P93k+12.4k & 72.01       \\
      \bottomrule
    \end{tabular}}
  \caption{Performance of direct pre-training with less data.}
  \label{less}
\end{table}

\section{Conclusion}
In this paper, We explored to setup a fast and efficient scratch training pipeline for object detection. Extensive experiments that proved that our proposed direct pre-training significantly improve the efficiency of training from scratch. And with more and more massive scale detection datasets like Open Images \cite{OpenImages}, Objects365 \cite{Objects365} recently introduced, we need to develope more efficient training strategies to tackle these cases. We hope our work will help community to move forward.

\section{Acknowledgements}
This work was supported by the National Natural Science Foundation of China (Grant No.62088101, 61673341), the Project of State Key Laboratory of Industrial Control Technology, Zhejiang University, China (No.ICT2021A10), the Fundamental Research Funds for the Central Universities (No.2021FZZX003-01-06), National Key R\&D Program of China (2016YFD0200701-3), Double First Class University Plan (CN).

\medskip

{\small
  \bibliographystyle{ieeetr}
  \bibliography{egbib}
}

\newpage
\section{Appendix}

\subsection{More Details of Experiments}

In this section, we provide more details of our experiments.

\tb{1. Normalization performance for Montage training.} For fair comparison Montage training\cite{Montage}, we use normalized Montage training. Specifically, for mAP \cite{Montage} reported relatively improve (+0.2mAP) \vs their ImageNet baseline, we add this to our ImageNet baseline(38.2mAP). And for time comparison, we use our ImageNet training baseline 9.05 GPU days multiplied 1/4(paper reported results) as normalized total training time.

\tb{2. Swin transformer training.} We following original paper\cite{Swin} training setting, which means we using AdamW\cite{AdamW} and mixed precision training\cite{MixedPrecision} in the pre-training phase. And mixed precision use apex implementation.\footnote{\href{https://github.com/NVIDIA/apex}{https://github.com/NVIDIA/apex}} We use typical (1333, 800) resizing strategy instead of multi-scale training strategy of the paper for consistency with above experiments.

\tb{3. VOC dataset training.} We following MMDetection\cite{MMDetection} setting for PASCAL VOC dataset, Fine-tuned for 4 epochs(12.4k iterations). We validate that direct pre-training 93k iterations and longer schedule still not catches up ImageNet pre-training.

\subsection{Extended Effective Comparison}

In this section, we provide detailed comparison for direct pre-training and DropBlock\cite{DropBlock} style training. It shares similarity with direct pre-training that it also use ImageNet style resizing strategy. Differently, direct pre-training resizes to (448, 448) and then fine-tuning with (1333, 800) while DropBlock style training whole process with images resized to (640, 640). Besides, it need large batch size(batch size=8) and SyncBN which means it need more memory and training time. We training with DropBlock style training on an 8*RTX 6000 GPU(24G memory) server for comparison, which is slightly faster than 2080Ti used in above experiments. results shows in table \ref{extend}. From the results we can see that direct pre-training still has advances over memory consumption, training time and accuracy.

\begin{table}[H]
  \centering
  \setlength{\tabcolsep}{3mm}{
    \begin{tabular}{c|ccc|cc}
      \toprule
      Method    & Schedule & Memory     & Cost(GPU days)        & $AP^{bbox}$ & $AP^{mask}$ \\
      \hline
      Direct    & P2x+1x   & \tb{8.06G} & \tb{2.07}(RTX 2080Ti) & \tb{41.0}   & \tb{36.6}   \\
      DropBlock & 3x       & 11.25G     & 3.32(RTX 6000)        & 39.8        & 35.5        \\
      \bottomrule
    \end{tabular}}
  \caption{Comparison with DropBlock style training.}
  \label{extend}
\end{table}

\end{document}